\documentclass[runningheads]{llncs}

\usepackage{makeidx}  
\usepackage{url}
\usepackage{color}
\usepackage{longtable}
\usepackage{multirow}
\usepackage{graphicx}
\usepackage{amssymb, amsmath, bm}
\usepackage{mathtools}
\usepackage{mathrsfs}
\usepackage{booktabs}
\usepackage{cite}
\usepackage{dsfont}
\usepackage[colorlinks,linkcolor=blue]{hyperref}
\usepackage{makecell}
\usepackage[noend]{algpseudocode}
\usepackage{algorithmicx,algorithm}
\usepackage{graphicx}

\usepackage[misc]{ifsym} 
\usepackage{bbding}
\begin{document}
	
\title{Online Reflective Learning for \\Robust Medical Image Segmentation}
\titlerunning{Online Reflective Learning for Robust Medical Image Segmentation}

\author{Yuhao Huang\inst{1,2,3}\thanks{Yuhao Huang and Xin Yang contribute equally to this work.}, Xin Yang\inst{1,2,3\star} \and Xiaoqiong Huang\inst{1,2,3} \and Jiamin Liang\inst{1,2,3} \and Xinrui Zhou\inst{1,2,3} \and Cheng Chen\inst{4} \and Haoran Dou\inst{5} \and Xindi Hu\inst{6} \and Yan Cao\inst{1,2,3} \and Dong Ni\inst{1,2,3}\textsuperscript{(\Letter)}} 


\institute{
\textsuperscript{$1$}National-Regional Key Technology Engineering Laboratory for Medical Ultrasound, School of Biomedical Engineering, Health Science Center, Shenzhen University, China\\
\email{nidong@szu.edu.cn} \\
\textsuperscript{$2$}Medical Ultrasound Image Computing (MUSIC) Lab, Shenzhen University, China\\
\textsuperscript{$3$}Marshall Laboratory of Biomedical Engineering, Shenzhen University, China\\
\textsuperscript{$4$}Department of Computer Science and Engineering, The Chinese University of Hong Kong, China\\
\textsuperscript{$5$}Centre for Computational Imaging and Simulation Technologies in Biomedicine (CISTIB), University of Leeds, UK\\ 
\textsuperscript{$6$}Shenzhen RayShape Medical Technology Co., Ltd, China\\}


\authorrunning{Huang et al.}
\maketitle              
\begin{abstract}
Deep segmentation models often face the failure risks when the testing image presents unseen distributions. 
Improving model robustness against these risks is crucial for the large-scale clinical application of deep models. 
In this study, inspired by human learning cycle, we propose a novel online reflective learning framework (\textit{RefSeg}) to improve segmentation robustness.
Based on the reflection-on-action conception, our RefSeg firstly drives the deep model to take action to obtain semantic segmentation. Then, RefSeg triggers the model to reflect itself. Because making deep models realize their segmentation failures during testing is challenging, RefSeg synthesizes a realistic proxy image from the semantic mask to help deep models build intuitive and effective reflections. This proxy translates and emphasizes the segmentation flaws. By maximizing the structural similarity between the raw input and the proxy, the reflection-on-action loop is closed with segmentation robustness improved. RefSeg runs in the testing phase and is general for segmentation models. 
Extensive validation on three medical image segmentation tasks with a public cardiac MR dataset and two in-house large ultrasound datasets show that our RefSeg remarkably improves model robustness and reports state-of-the-art performance over strong competitors.
\end{abstract}
	
\section{Introduction}
\label{sec:intro}
\begin{figure*}[t]
	\centering
	\includegraphics[width=0.69\linewidth]{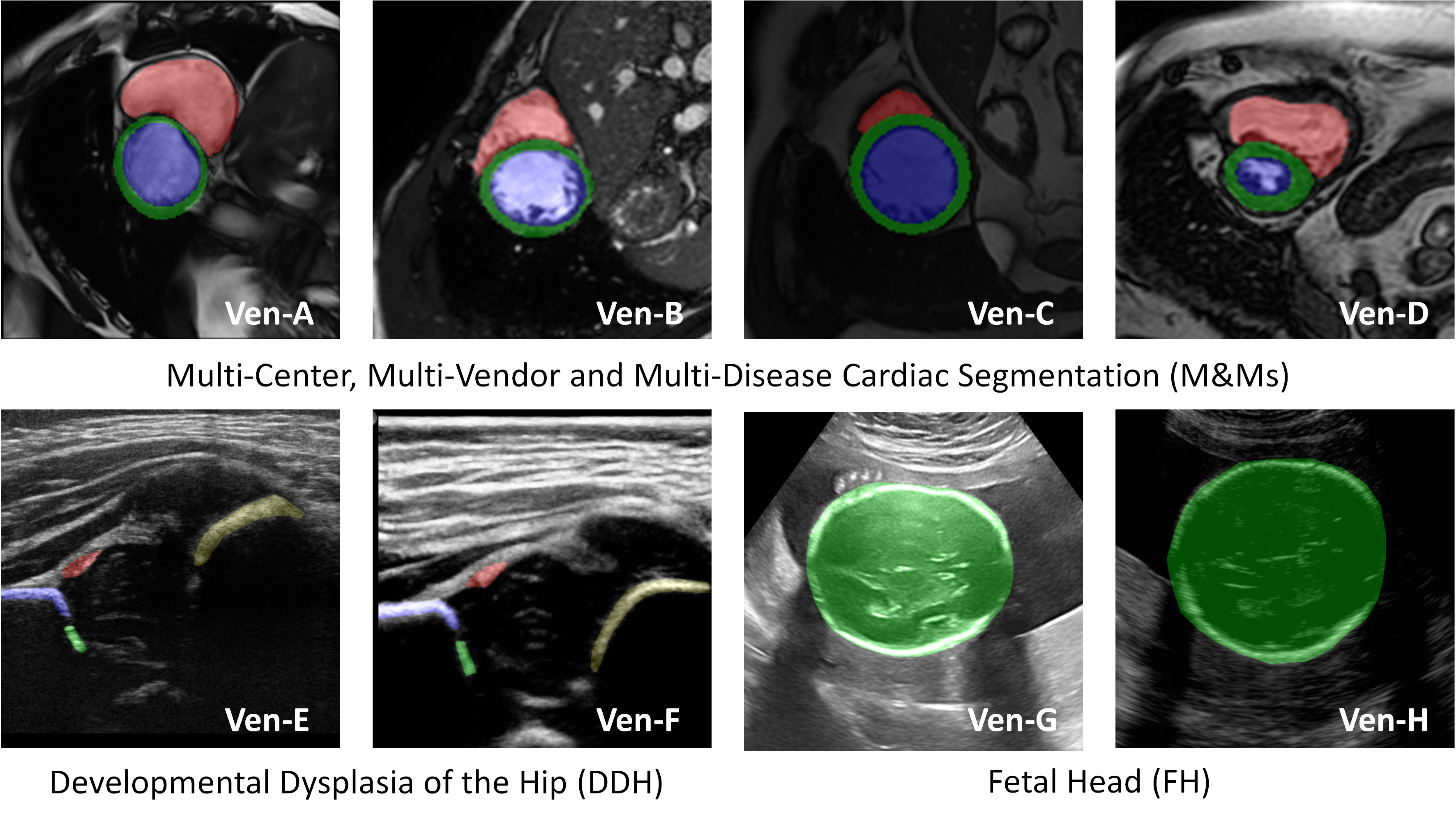}
	\caption{Visualization of datasets from different vendors (Ven-A to Ven-H). Eight structures to be segmented include (1) M\&Ms: left ventricle (LV, blue), left ventricular myocardium (MYO, green) and right ventricle (RV, red); (2) DDH: ilium (blue), lower limb (green), labrum (red) and co\_junction (yellow); (3) FH: head (green).}
	\label{fig:intro}
\end{figure*}

Deep learning has achieved great success in medical image segmentation~\cite{shen2017deep}.
However, deep models heavily depend on the training data and can easily suffer from serious performance degradation \cite{dou2019domain,liu2021style} when deploying to unseen test data with distribution shift, e.g., images acquired from different centers, devices, operators or acquisition protocols (Fig.~\ref{fig:intro}). 
This risk greatly threats the reliable deployment of deep segmentation models in critical medical applications. 
To achieve robust segmentation, a plethora of methods have been proposed to address the risk and can be roughly categorized into three mainstreams.

\textbf{Domain Adaptation (DA)}~\cite{guan2021domain} typically aims to align distributions by using both source domain labeled data and target domain data. Most DA methods explore image or feature alignment and invariant feature extraction across domains. Regarding the label availability of target domain, DA can be summarized as supervised~\cite{motiian2017unified,kushibar2019supervised}, semi-supervised~\cite{madani2018semi} and unsupervised DA (UDA)~\cite{huo2018synseg,zhang2020collaborative,chen2021source}. Though UDA obtains promising performance on cross-domain medical images and does not require annotations of target domains, obtaining adequate amount of target images in advance can be tough for clinical practice. \par

\textbf{Domain Generalization (DG)}~\cite{bergamo2010exploiting} only relies on source domains to train robust models that can directly generalize to different unseen domains. Though appealing, most existing methods need more than one source domain for training~\cite{shankar2018generalizing}. 
The alternatives are domain randomization~\cite{volpi2018generalizing,billot2021synthseg} and adversarial learning~\cite{tobin2017domain} to systematically augment training samples to cover possible cases. 
Recently, Liu et al.~\cite{liu2021style} proposed a style-transfer based curriculum learning to enhance segmentation generalisation.
Meta-learning based methods~\cite{liu2020shape,liu2021semi} were introduced to generalize segmentation models to unseen domains.
Although effective, these methods often suffer from the problems of difficulties in controlling data generation, style image selection, and model over-fitting. \par

\textbf{Test-time Adaptation (TTA)} is an emerging topic that adapts a pre-trained model to each inference image at test time. Among the studies, test-time augmentation has preferred efficacy~\cite{moshkov2020test,liu2021style}. Although it is easy to implement, it cannot tightly mimic the unseen domains with limited augmentation combinations. Recently, self-supervised learning was introduced to remove the domain shift during testing~\cite{yang2018generalizing}. 
Karani et al. \cite{karani2021test} proposed an autoencoder to extract the anatomical prior for guidance. 
He et al. \cite{he2021autoencoder} introduced a set of adaptors to minimize the reconstruction errors. These methods made great efforts in exploring self-supervision signals for effective test-time adaptation, but these signals are still not accurate and strong enough to drive the deep model to clearly realize its segmentation failures for correction. \par

In this work, we propose a novel online Reflective Learning framework, named \textit{RefSeg}, to quickly achieve robust segmentation at test time. RefSeg is inspired by human's intuition of how to reflect and identify the segmentation failure to refine. Specifically, based on the predicted semantic mask, humans would reflect and reconstruct an imaginary proxy image. 
If the segmentation is accurate, this proxy would present similar structures to the original input image, otherwise, this proxy would have remarkable flaws.
This discrepancy drives the iterative reflection-on-action and helps humans to robustly segment images from unseen domains. Following this reflection-on-action conception, RefSeg drives the deep model to act for semantic segmentation. Then, RefSeg would encourage the deep model to reflect itself and identify segmentation flaws by synthesizing a realistic proxy image from the semantic mask. By maximizing the structural similarity between the raw input and the proxy, the deep model would be able to remedy its weakness and gradually converge towards a plausible segmentation. \par

Our RefSeg is an important extension of self-supervised learning. It significantly improves the efficacy of supervision signal by reflecting on the mask.
Notably, the source domain information is only used for offline model training in RefSeg. During inference, RefSeg directly conducts online adaptation of the trained model to the unseen testing image by iteratively reflecting and refining the segmentation. 
We conduct extensive experiments on three segmentation tasks with public M\&Ms Challenge~\cite{campello2021multi} and two in-house large US datasets.
We demonstrate that RefSeg is general in reducing the failure risk of deep segmentation models, remarkably improves the model robustness, achieves state-of-the-art results, and even approaches the supervised training upper bound on some tasks. \par

\section{Methodology}
\label{sec:method}
\begin{figure*}[!t]
	\centering
	\includegraphics[width=0.83\linewidth]{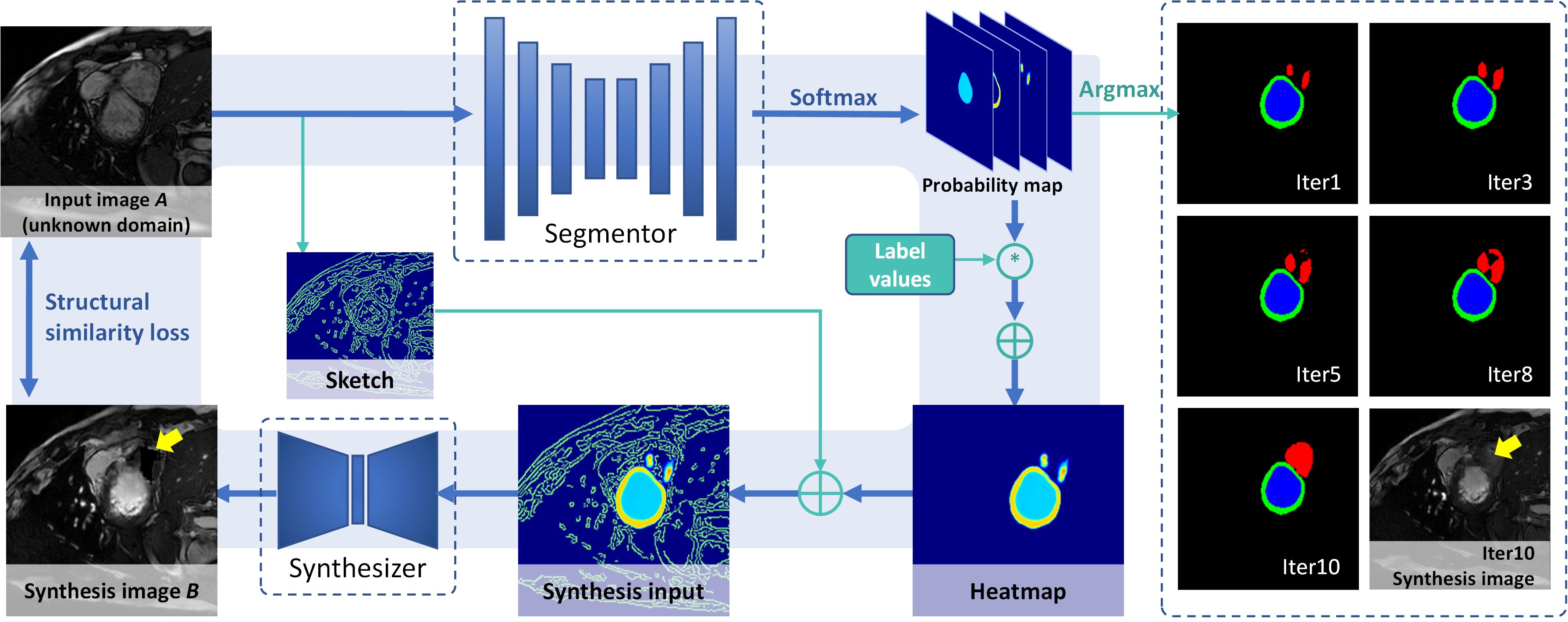}
	\caption{The workflow of our proposed Reflective Learning framework (RefSeg).}
	\label{fig:framework}
\end{figure*}
Fig.~\ref{fig:framework} shows the workflow of RefSeg. For each unknown-domain testing image, 
RefSeg drives the trained segmentor to produce segmentation heatmap. 
Then, a synthesizer leads the reflection process by generating a realistic proxy image based on the heatmap and the auxiliary sketch input. Our introduced structural similarity loss then measures the segmentation failure. Both the segmentor and synthesizer are then fine-tuned under this supervision for robust segmentation. \par

\subsubsection{Segmentor for Differentiable Learning.}
As shown in Fig.~\ref{fig:framework}, RefSeg is a general framework for $k$-class segmentation ($k \geq 2$, with background). 
Commonly, performing the non-differentiable \textit{argmax} on the probability map can obtain the segmentation results.
Here, to connect the gradient flow of the segmentation action path and synthesis reflection path for differentiable online learning cycle, we multiply probability map of each channel $p_k$ by the corresponding ground truth label intensity value $g_k$ and add them together (i.e., ${\textstyle \sum}(p_k*g_k)$).
Then, the $k$-class heatmap $H_{att}$ with gradient can be produced. 

\subsubsection{Synthesizer for Informative Proxy.}
The source domain-trained segmentor suffers from failures in testing unseen-domain images. 
However, it is hard for the segmentor to reflect on its segmentation result and identify the failure. 
RefSeg provides an easy and intuitive reflection scheme through a synthesizer (see Fig.~\ref{fig:framework}). 
It can generate a realistic proxy image, which translates and emphasizes the segmentation flaws. When the segmentation fails, it will be reflected in the synthesis image (yellow arrows in Fig.~\ref{fig:framework}) in details. We believe that this reflection can provide informative and direct cues for failure risk evaluation. Specifically, inspired by~\cite{liang2022sketch}, we adopt the Generative Adversarial Network (GAN) to synthesize proxy based on the heatmap and the sketches (like canny edge) obtained from the original input image. The heatmap indicates the target structures, while the sketch provides important support for realistic details. 
Same as the segmentor, the synthesizer is trained and validated on source domain.
\par

\subsubsection{Structural Similarity for Failure Evaluation.}
RefSeg reflects on the segmentation failure by quantifying the similarity between the input and synthesized proxy images. It is worth noting that, because the synthesizer is trained on the source domain, it can only translate the heatmap into the source domain by default. Therefore, as Fig.~\ref{fig:framework} shows, the input image comes from an unknown domain, while the synthesis image locates in the source domain. 
They naturally have unknown domain shifts. 
Regarding this challenge, inspired by multi-modal registration, we propose the structural similarity as measurement, rather than the pixel-level content similarities (like \textit{L1} loss). Specifically, our similarity loss has the following main complementary components. \par

\textbf{1) Normalized Cross-Correlation Loss (NCC).}
Invented to measure the similarity between an image and template, NCC has been extensively employed in image registration tasks~\cite{sarvaiya2009image}. The range of NCC is $[-1,1]$, with $-1$ and $1$ indicating perfect negative and positive correlation, respectively. Given the input image \textit{A} and proxy image \textit{B}, they both are divided into $N$ sub-images with size $n$ ($n$=9 in our work). By calculating the CC of two corresponding sub-images ($A(x, y)$ and $B(x, y)$) and averaging $N$ pairs, NCC loss can be formulated as:

\begin{small}
\begin{align} L_{ncc}(A,B) & = 1 - \frac{1}{N} \sum_{x, y} \frac{1}{\sigma_{A} \sigma_{B}}\left(A(x, y)-\mu_{A}\right)\left(B(x, y)-\mu_{B}\right), \end{align}
\end{small}
where $\mu_{\cdot }$ and $\sigma_{\cdot }$ are the mean and standard deviation of each sub-image. \par

\textbf{2) Mutual Information Loss (MI).} 
MI comes from the information theory, measuring the shared information between two images (\textit{A} and \textit{B})~\cite{wells1996multi}. Different from the NCC that usually adopted in single-modal registration, MI is preferred in describing the common features shared among multiple modalities. The higher MI ($MI \geq 0$) means more information are shared between two images. Mathematically, MI can be defined as:
\begin{align}
\label{equ: MI}
\begin{small}
MI(A,B) = \sum_{a\in \mathcal{A} } \sum_ {b \in \mathcal{B} } p(a,b)log\frac{p(a,b)}{p(a)p(b)},
\end{small}
\end{align}
where $a$ and $b$ represent the intensity values of A and B, respectively. $p(\cdot)$ denotes the probability, relying on histogram-based calculation of pixel intensities, and thus is non-differentiable. For approximating MI in a differentiable way, we exploit the \textit{Parzen window} formulation with \textit{Gaussian} kernel to estimate $\tilde{MI}$ (refer to ~\cite{xu2008parzen}). Then, the MI loss can be defined as $L_{mi}(A,B) = -\tilde{MI}(A,B)$.

\textbf{3) Attention on Target Structures.}
In order to emphasize the target structures for segmentation in the structural similarity loss, we further propose to leverage the normalized heatmap $H'_{att}=H_{att}/255$ as the attention matrix to indicate the desired segmentation areas. 
Hence, the final image input to the similarity loss is $\hat{A} = (\textit{\textbf{I}}+H'_{att})\cdot A$, $\hat{B} = (\textit{\textbf{I}}+H'_{att})\cdot B$. $\textit{\textbf{I}}$ is the matrix with all the elements equal to 1. Therefore, our online RefSeg iteratively minimizes the final similarity loss on each testing image and remedies its segmentation failures for robustness as follows: $Loss = L_{ncc}(\hat{A},\hat{B}) + L_{mi}(\hat{A},\hat{B})$.
\section{Experimental Results}
\label{sec:Experimental Results}
\subsubsection{Materials and Implementations.}
We validated RefSeg on three segmentation tasks (Fig. \ref{fig:intro}).
For \textbf{Cardiac segmentation}, we employ cardiac MR images from M\&Ms Challenge~\cite{campello2021multi}, which aims to segment cardiac substructures including \textit{LV}, \textit{MYO}, and \textit{RV}. 
The dataset consists of volumes acquired from four different vendors. 
We follow the experimental setting of the challenge to train on images from vendor A\&B and test on images from all vendors A to D.
For \textbf{DDH segmentation}, the aim is to segment \textit{ilium}, \textit{lower limb}, \textit{labrum} and \textit{co\_junction}. 
We use one in-house dataset with images collected from two vendors E and F.
For \textbf{FH segmentation}, the only structure to be segmented is \textit{head}.
We use another in-house dataset with images acquired from two vendors G and H. 
The two private datasets were manually annotated by experts using the Pair annotation software package~\cite{liang2022sketch}.
We conduct bidirectional experiments for vendors on DDH and FH datasets for extensive evaluations, denoted as E2F, F2E, G2H and H2G. Dataset split information are listed in Table~\ref{tab:dataset}. For each split, the segmentor and synthesizer used the same training, validation and testing set. 

\begin{table*}[!t]
    \centering
    \scriptsize
    \begin{center}
    \caption{Datasets split for the training, validation and testing set of each experimental group (volumes/cases (slices/images)). The M\&Ms dataset was split following the Challenge. The other two in-house datasets were split randomly at patient/case level.}
    \setlength{\tabcolsep}{3.0mm}{
    \begin{tabular}{c|cc|cc|cc}
    \toprule[1pt]
    & \multicolumn{2}{c|}{Training}  &\multicolumn{2}{c|}{Validation} & \multicolumn{2}{c}{Testing}                           \\ 
    \hline
    \multirow{4}{*}{M\&Ms}    & \multicolumn{1}{c|}{\multirow{2}{*}{A}} & \multicolumn{1}{c|}{\multirow{2}{*}{75 (1738)}} & \multicolumn{1}{c|}{\multirow{2}{*}{A}} & \multicolumn{1}{c|}{\multirow{2}{*}{4 (98)}}   & \multicolumn{1}{c|}{A} & \multicolumn{1}{c}{16 (360)}  \\ 
    & \multicolumn{1}{c|}{} & \multicolumn{1}{c|}{} & \multicolumn{1}{c|}{}                   & \multicolumn{1}{c|}{} & \multicolumn{1}{c|}{B} & \multicolumn{1}{c}{40 (888)}  \\ 
    & \multicolumn{1}{c|}{\multirow{2}{*}{B}} & \multicolumn{1}{c|}{\multirow{2}{*}{75 (1546)}} & \multicolumn{1}{c|}{\multirow{2}{*}{B}} & \multicolumn{1}{c|}{\multirow{2}{*}{10 (208)}} & \multicolumn{1}{c|}{C} & \multicolumn{1}{c}{40 (974)}  \\ 
    & \multicolumn{1}{c|}{} & \multicolumn{1}{c|}{}  & \multicolumn{1}{c|}{} & \multicolumn{1}{c|}{} & \multicolumn{1}{c|}{D} & \multicolumn{1}{c}{40 (1014)} \\ 
    \hline
    \multicolumn{1}{c|}{DDH (E2F)}   & \multicolumn{1}{c|}{E} & 214 (402) & \multicolumn{1}{c|}{E} & 98 (180) & \multicolumn{1}{c|}{F} & 320 (649) \\ 
    \hline
    \multicolumn{1}{c|}{DDH (F2E)}   & \multicolumn{1}{c|}{F} & 220 (249) & \multicolumn{1}{c|}{F}  & 100 (200)  & \multicolumn{1}{c|}{E} & 312 (582) \\ 
    \hline
    \multicolumn{1}{c|}{FH (G2H)} & \multicolumn{1}{l|}{G} & 75 (300) & \multicolumn{1}{c|}{G} & 31 (124) & \multicolumn{1}{l|}{H} & 199 (796) \\ 
    \hline
    \multicolumn{1}{c|}{FH (H2G)} & \multicolumn{1}{l|}{H}  & 150 (600)  & \multicolumn{1}{c|}{H} & 49 (196)  & \multicolumn{1}{l|}{G} & 106 (424) \\ 
    \toprule[1pt]
    \end{tabular}}
\label{tab:dataset}
\end{center}
\end{table*}

We implemented RefSeg in Pytorch, using an NVIDIA TITAN 2080 GPU. All images were resized to 256$\times$256. We chose Attention U-net~\cite{oktay2018attention} as a typical segmentor and trained it with Adam optimizer, learning rate (\textit{lr})=1e-3 and batch size=8. During offline training, segmentor was supervised by cross-entropy loss. The synthesizer was trained following~\cite{liang2022sketch}, using Adam optimizer with batch size=4. The \textit{lr} for generator and discriminator are 1e-3 and 1e-4, respectively. Training epochs for segmentor and synthesizer are 100 and 400. We selected models with the best performance on validation sets to work with RefSeg. During the online testing, the \textit{lr} for segmentor are 1e-4, 1e-5 and 1e-5 for M\&Ms, DDH and FH datasets, respectively. For synthesizers, the optimal \textit{lr} are 1e-4 for all datasets. RefSeg iterates for 10 steps in all experiments. \par

\subsubsection{Quantitative and Qualitative Analysis.}
We evaluated the segmentation performance using Dice similarity coefficient (DICE) and Hausdorff distance (HD) for all the experiments. For M\&Ms dataset, we added the Challenge criteria (\textit{DICE score} and \textit{HD score}) and \textit{min-max (MM) score} ranking method for fair comparisons~\cite{campello2021multi}. 
For each metric, the average results of substructures are reported.
Table~\ref{tab:mms} compares RefSeg with other seven methods on M\&Ms, including Attention U-net~\cite{oktay2018attention} (baseline), a recent robust segmentation method SCL~\cite{liu2021style}, two state-of-the-art TTA methods DAE~\cite{karani2021test} and SDAN~\cite{he2021autoencoder}, and the top-3 methods on the Challenge leaderboard (i.e., P1-P3). Note that since we exactly follow the experimental setting of the M\&Ms challenge, methods P1-P3 based on the heavy nnUNet~\cite{isensee2021nnu} are strong comparisons.
It can be observed that our proposed RefSeg improves over baseline by 4.81\% and 1.95\textit{mm} in terms of the average DICE and HD score. Based on the \textit{MM score}, RefSeg achieves the best performance among strong competitions. Moreover, RefSeg only requires 1.72s (0.172s/step). It is much faster than the P1 (26s) and P2 (4.8s). \par

\begin{table*}[!t]
	\centering
	\scriptsize
	\begin{center}
		\caption{Comparisons on M\&Ms Challenge (mean(std)) (DICE$\uparrow$:~\%, HD$\downarrow$:~mm). }
			\setlength{\tabcolsep}{0.3mm}{
			\begin{tabular}{l|cc|cc|cc|cc|ccc}
				\toprule[1pt]
				\multirow{2}{*}{\textbf{Methods}} & \multicolumn{2}{c|}{Vendor A} &
				\multicolumn{2}{c|}{Vendor B} & \multicolumn{2}{c|}{Vendor C} & \multicolumn{2}{c|}{Vendor D} &
				\multirow{2}{*}{\shortstack{\textbf{DICE}\\\textbf{Score}}}
				&
				\multirow{2}{*}{\shortstack{\textbf{HD}\\\textbf{Score}}} 
				
				& \multirow{2}{*}{\shortstack{\textbf{MM}\\\textbf{Score}}} 
				\\
				\cline{2-9}
				
				\specialrule{0em}{0.8pt}{0.5pt}
				& DICE  & HD  & DICE  & HD & DICE & HD & DICE & HD  \\
				
				\hline
			
			    \multirow{2}{*}{SDAN~\cite{he2021autoencoder}} 
				& 86.59 & 12.04 & 85.23 & 11.63
				& 77.23 & 14.06 & 82.68 & 11.64 & \multirow{2}{*}{81.94}& \multirow{2}{*}{12.51}& \multirow{2}{*}{24.74} \\
				& (4.51) & (3.88) & (7.91) & (5.59) & (17.78) & (8.34) & (6.48) & (6.26) \\
				\hline
				
				\multirow{2}{*}{Baseline~\cite{oktay2018attention}} 
				& 87.19 & 11.55  & 87.05 & 9.78 
				& 79.52 & 13.53 & 83.82 & 9.80 & \multirow{2}{*}{83.49}& \multirow{2}{*}{11.33}& \multirow{2}{*}{47.03} \\
				& (5.10) & (3.48) & (5.36) & (3.15) & (14.46) & (11.92) & (5.10) & (4.11) \\
				
				\hline
				
				\multirow{2}{*}{DAE~\cite{karani2021test}} & 87.03 & 11.77 & 86.74 & 9.95 & 78.97 & 12.18 & 84.10 & 10.11 & \multirow{2}{*}{83.32} & \multirow{2}{*}{11.05}& \multirow{2}{*}{48.46} \\
				& (4.61) & (3.43) & (5.88) & (3.96) & (16.13) & (6.10) & (5.58) & (4.36)  \\	
				\hline

				\multirow{2}{*}{P3~\cite{ma2020histogram}} & 87.85 & 12.59 & 88.68 & 9.87 & 86.47 & 10.46 & 86.41 & 13.97 & \multirow{2}{*}{87.05}& \multirow{2}{*}{11.89}& \multirow{2}{*}{67.26}  \\
				& (4.25) &  (12.28) & (5.29) & (3.61) & (5.54) & (6.50) & (5.07) & (16.85)  \\
				
				\hline
				
				\multirow{2}{*}{P2~\cite{zhang2020semi}} & 88.39 & 12.46  & 89.16 & 9.80 & 86.96 & 9.55 & 86.68 & 13.43 & \multirow{2}{*}{87.47}& \multirow{2}{*}{11.37}& \multirow{2}{*}{75.20}  \\
				& (3.96) &  (12.24) & (5.10) & (3.37) & (4.66) & (3.29) & (4.96) & (14.85)  \\
				
				\hline
				
				\multirow{2}{*}{SCL~\cite{liu2021style}} & 88.42 & \textcolor{blue}{9.80} & 87.64 & 10.24 & 86.48 & 10.59 & 86.79 & 11.20 & \multirow{2}{*}{87.10} & \multirow{2}{*}{10.60}& \multirow{2}{*}{79.84} \\
				& (3.23) &  (3.53) & (4.69) & (2.98) & (6.00) & (5.19) & (3.67) & (4.89)  \\
				
				\hline
				
				\multirow{2}{*}{P1~\cite{full2020studying}} & 88.91 & 12.07 & \textcolor{blue}{89.28} & 9.48 & 87.63 & 9.47 & \textcolor{blue}{87.66} & 13.09 & \multirow{2}{*}{88.13}& \multirow{2}{*}{11.11}& \multirow{2}{*}{82.40}  \\
				& (4.19) &  (12.64) & (4.65) & (3.34) & (4.26) & (3.56) & (4.21) & (14.84)  \\
		
				\hline
				\hline
				
				\multirow{2}{*}{RefSeg-L1} 
				& 84.73 & 14.10 & 85.01 & 12.44 
				& 78.05 & 19.17 & 81.09 & 11.53 & 
				\multirow{2}{*}{81.33}& 
				\multirow{2}{*}{14.66}& \multirow{2}{*}{0.00}  \\
				& (6.92) & (6.72) & (5.78) & (5.01) 
				& (9.82) & (18.64) & (5.27) & (4.51) \\
				
				\hline
				
				\multirow{2}{*}{RefSeg-N} 
				& 88.27 & 10.37 & 87.92 & 9.84
				& 81.97 & 12.95 & 85.37 & 9.71 & 
				\multirow{2}{*}{85.15}& 
				\multirow{2}{*}{10.92}& 
				\multirow{2}{*}{62.82}  \\
				& (5.02) & (4.85) & (4.63) & (3.47) 
				& (9.85) & (12.62) & (3.76) & (4.02) \\
				
				\hline
				
				\multirow{2}{*}{RefSeg-M}
				& 88.65 & 11.21 & 88.02 & 9.98 
				& 83.08 & 13.70 & 86.74 & \textcolor{blue}{8.05} & 
				\multirow{2}{*}{86.05}& 
				\multirow{2}{*}{10.78}& 
				\multirow{2}{*}{70.60}  \\
				& (3.55) & (3.49) & (3.09) & (2.84) 
				& (8.68) & (15.96) & (3.42) & (3.61) \\
				
				\hline
				
				\multirow{2}{*}{RefSeg-NM}
				& \textcolor{blue}{89.02} & 10.32 & 87.91 & \textcolor{blue}{9.24} 
				& 84.68 & 11.85 & 87.21 & 9.05 & 
				\multirow{2}{*}{86.79}& 
				\multirow{2}{*}{10.23}& 
				\multirow{2}{*}{81.12}  \\
				& (3.41) & (2.41) & (3.85) & (2.44) 
				& (8.13) & (12.11) & (3.48) & (3.16) \\
				
				\hline
				\multirow{2}{*}{\textbf{RefSeg}}
				& 88.89 & 10.65 & 88.66 & 9.31 
    			& \textcolor{blue}{88.97}  & \textcolor{blue}{9.34}  & 87.15 & 8.83 & 
				\multirow{2}{*}{\textcolor{blue}{88.30}}& 
				\multirow{2}{*}{\textcolor{blue}{9.38}}& 
				\multirow{2}{*}{\textcolor{blue}{100.00}}  \\
				& (4.50) & (3.36) & (4.53) & (2.81) 
				& (9.54) & (6.18) & (3.72) & (4.38) \\
				\toprule[1pt]
		\end{tabular}}
		\label{tab:mms}
	\end{center}
\end{table*}

\begin{table*}[!t]
	\centering
	\scriptsize
	\begin{center}
		\caption{Comparisons on DDH and FH datasets (mean(std)) (DICE:$\uparrow$~\%, HD$\downarrow$:~pixel).}
		\setlength{\tabcolsep}{0.7mm}{
			\begin{tabular}{l|cc|cc|cc|cc}
				\toprule[1pt]
				\multirow{2}{*}{\textbf{Methods}} & \multicolumn{2}{c|}{DDH: E2F} & \multicolumn{2}{c|}{DDH: F2E} & \multicolumn{2}{c|}{FH: G2H} & \multicolumn{2}{c}{FH: H2G} \\
				\cline{2-9}
				\specialrule{0em}{0.8pt}{0.5pt}
				& DICE  & HD  & DICE  & HD & DICE  & HD & DICE & HD  \\
				\hline
				
				\multirow{2}{*}{Upper-bound} 
				& 88.21 & 11.76 & 87.64 & 11.25
				& 98.22 & 3.51 & 98.45 & 3.21 \\
				& (7.21) & (10.63) & (8.21) & (9.88) 
				& (1.24) & (6.37) & (3.18) & (2.74) \\
				\hline
				\hline
				\multirow{2}{*}{Baseline~\cite{oktay2018attention}} 
				& 83.47 & 13.51 & 72.54 & 15.99 
				& 90.17 & 6.97 & 97.53 & 5.43 \\
				& (5.91) & (10.74) & (16.07) & (16.98) 
				& (16.53) & (10.79) & (1.76) & (4.23) \\
				\hline
				
				\multirow{2}{*}{DAE~\cite{karani2021test}} 
				& 83.35 & 7.04 & 71.48 & 20.47 
				& 73.04 & 14.88 & 97.61 & 3.43  \\
				& (5.57) & (6.17) & (14.20) & (18.84) 
				& (36.69) & (25.51) & (1.22) & (1.92)   \\
				\hline
                
				\multirow{2}{*}{SDAN~\cite{he2021autoencoder}}
				& 83.10 & 8.23 & 67.01 & 18.86
				& 91.07 & 11.00 & 97.47 & 3.94  \\
				& (5.76) & (8.89) & (24.22) & (9.64)
				& (12.99) & (14.67) & (1.77) & (4.23) \\
				\hline
				
				\multirow{2}{*}{SCL~\cite{liu2021style}}
				& 84.76 & \textcolor{blue}{5.75} & 74.84 & 13.92
				& 79.79 & 11.89 & 98.15 & 2.84  \\
				& (5.11) & (2.88) & (14.58) & (8.05) 
				& (31.19) & (18.59)  & (0.89) & (1.55) \\
				\hline
				
				\multirow{2}{*}{nnUNet~\cite{isensee2021nnu}}
				& 85.38 & 5.91 & 80.39 & \textcolor{blue}{10.62} 
				& 89.43 & 13.71 & 98.07 & \textcolor{blue}{2.63} \\
				& (4.34) & (4.77) & (8.03) & (10.35) 
				& (24.72) & (28.72) & (0.67) & (0.93)  \\
				\hline
				
				\multirow{2}{*}{\textbf{RefSeg}} 
				& \textcolor{blue}{87.86} & 6.91 & \textcolor{blue}{81.56} & 14.34
				& \textcolor{blue}{97.31} & \textcolor{blue}{3.74} & \textcolor{blue}{98.26} & 3.07 \\
				& (3.01) & (9.24) & (8.52) & (14.57) 
				& (4.58) & (5.94) & (0.81) & (4.35) \\
				\toprule[1pt]
		\end{tabular}}
		\label{tab:ddhhc}
	\end{center}
\end{table*}

\begin{figure*}[!t]
	\centering
	\includegraphics[width=0.85\linewidth]{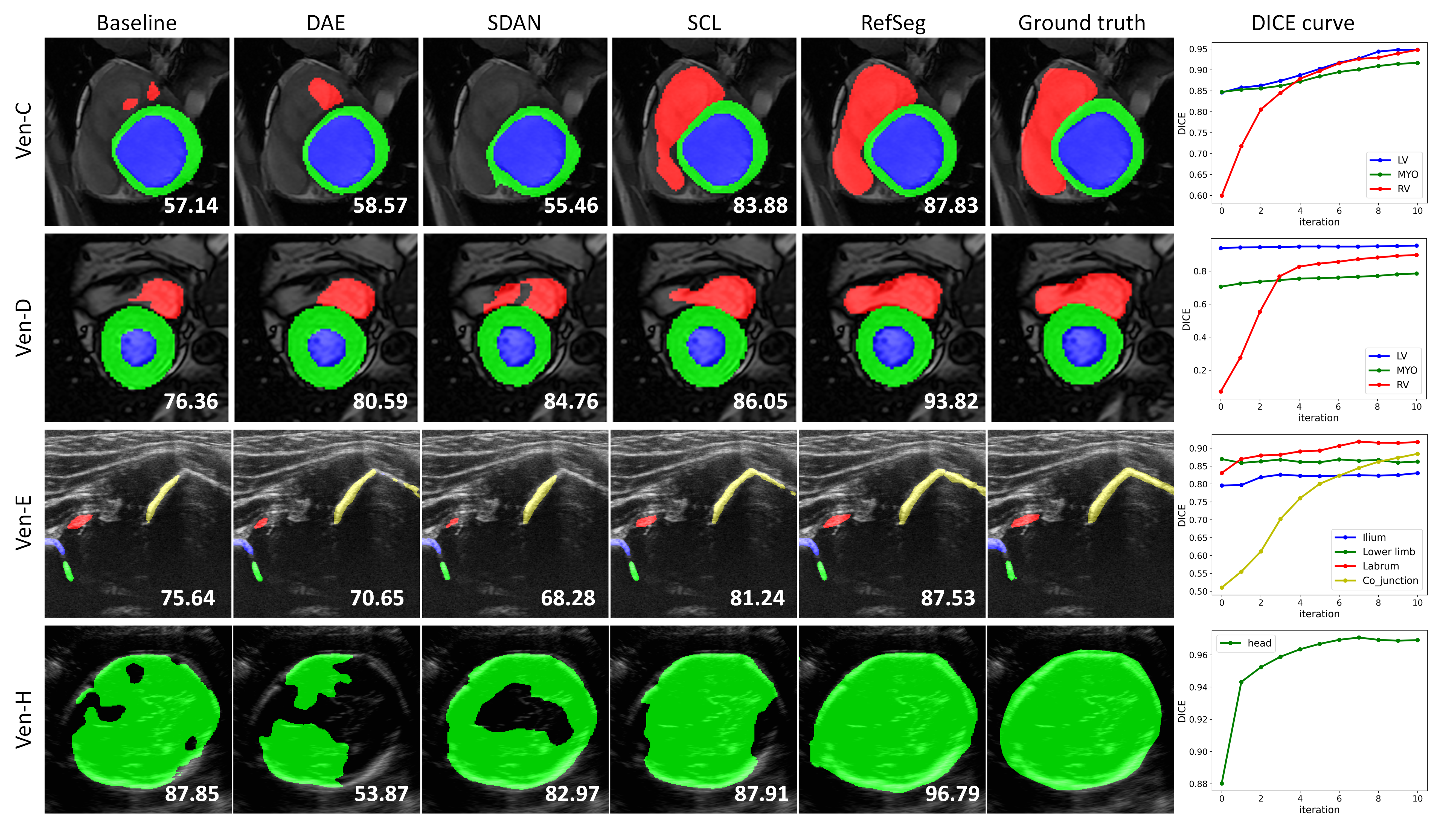}
	\caption{Typical results of RefSeg. DICE values are shown at the right image corner.}
	\label{fig:result}
\end{figure*}

\begin{figure*}[!t]
	\centering
	\includegraphics[width=0.9\linewidth]{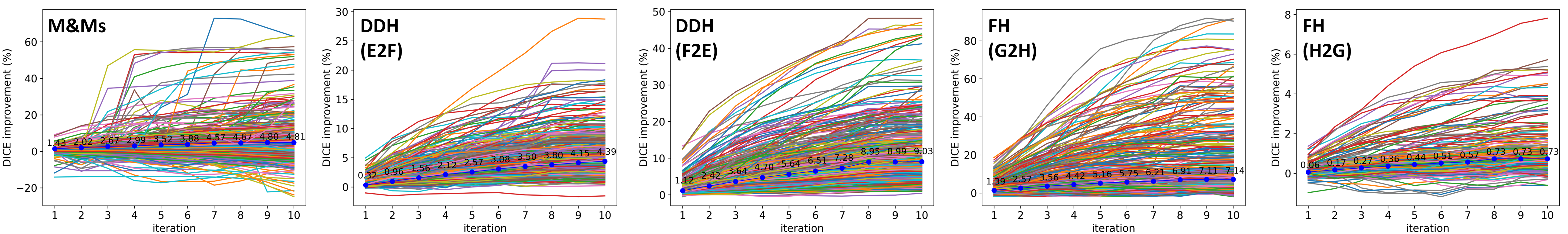}
	\caption{DICE improvement curves of our RefSeg. One curve for one testing sample. The blue dot curve indicates the average DICE over all testing samples.}
	\label{fig:online_curve}
\end{figure*}

We conduct ablation experiments on M\&Ms dataset (last 5 rows in Table~\ref{tab:mms}) to analyze the contribution of each key design for the important component of structural similarity measurement in our method. 
We compare our RefSeg with four other variants, including similarity measurement with only L1 loss (RefSeg-L1), NCC loss (RefSeg-N), or MI loss (RefSeg-M), as well as that with both NCC and MI losses but without the heatmap attention (RefSeg-NM).
Compared with \textit{baseline}, each of our proposed similarity losses and their combination can help improve the segmentation robustness, but RefSeg with L1 similarity loss is worse. It proves that maximizing the pixel-wise content similarity has adverse effects on RefSeg and our proposed structural similarity measurement captures important semantic information. We also note that the heatmap attention weight is effective in improving 1.51\% on the DICE score (RefSeg-NM vs. RefSeg). \par

Table~\ref{tab:ddhhc} reports the results of RefSeg and five competitors on DDH and FH datasets, including baseline~\cite{oktay2018attention}, DAE~\cite{karani2021test}, SDAN~\cite{he2021autoencoder}, SCL~\cite{liu2021style}, and nnUNet~\cite{isensee2021nnu}.
\textit{Upper-bound} represents the performance when training and testing on the target domain. 
Comparing \textit{Upper-bound} with~\textit{baseline}, the performance degradation can be observed clearly in DICE index. 
RefSeg can significantly boost the performance over baseline and achieved the highest DICE in all tests, approaching the \textit{Upper-bound} in E2F, G2H, and H2G.
Although the performance degradation in H2G is not as serious as other tests (about 0.92\% in DICE), our method can still achieve 0.73\% and 2.36 pixel improvement in DICE and HD. \par

Fig.~\ref{fig:result} visualizes the segmentation results of baseline, DAE, SDAN, SCL and RefSeg on three datasets. Column~\textit{baseline} shows the severe segmentation failures. 
Though DAE, SDAN and SCL can improve model performance in some cases, they may fail in others and even perform worse than the baseline.
From the DICE curves, we can see that RefSeg can significantly remedy the segmentation flaws iteratively and converges at high DICE. Fig.~\ref{fig:online_curve} shows DICE improvement curves of RefSeg on all five experiment groups. The rising trend of the averaged blue dot curve validates the efficacy of RefSeg. \par

\section{Conclusion}
\label{sec:conclusion}
In this study, we propose a novel online reflective learning framework (RefSeg) to help deep models recognize segmentation failures and remedy. RefSeg is general in improving the model robustness against unseen imaging factors. Extensive experiments on three large datasets validate that RefSeg is effective and efficient, achieving state-of-the-art results over \textit{nnUNet}. In the future, we will try to explore better similarity loss and extend the RefSeg to 3D segmentation tasks.
\subsubsection{Acknowledgement.} 
This work was supported by the grant from National Natural Science Foundation of China (Nos. 62171290, 62101343), Shenzhen-Hong Kong Joint Research Program (No. SGDX20201103095613036), and Shenzhen Science and Technology Innovations Committee (No. 20200812143441001).

\bibliographystyle{splncs04}
\bibliography{paper1267}
\end{document}